\documentclass[letterpaper, 10 pt, conference]{ieeeconf}  

\IEEEoverridecommandlockouts                              

\usepackage[table]{xcolor}

\usepackage{graphicx}
\usepackage{tikz}
\usetikzlibrary{decorations.pathreplacing}
\usepackage{amsmath}
\usepackage[outdir=./]{epstopdf}
\DeclareGraphicsExtensions{.eps,.jpg}
\usepackage{balance}
\usepackage{flushend}
\usepackage{float}
\usepackage{siunitx}
\usepackage{amsmath}
\usepackage{caption}
\usepackage{subcaption}

\usepackage{amssymb}
\usepackage{algorithm}
\usepackage{algorithmic}
\usepackage{tikz}
\usepackage{siunitx}
\usepackage{array}
\usepackage{bbold}

\newcolumntype{C}[1]{>{\centering\let\newline\\\arraybackslash\hspace{0pt}}m{#1}}

\usepackage{accents}  

\def \poses{X}
\def \landmarks{L}
\def \associations{D}
\def \measurements{Z}

\def \association{d}

\usepackage{accents}  

\def \poses{X}
\def \landmarks{L}
\def \associations{D}
\def \measurements{Z}

\def \association{d}

\title{\LARGE \bf
Open-Set Semantic Uncertainty Aware Metric-Semantic Graph Matching
}

\author{Kurran Singh$^1$ and John Leonard$^1$
\thanks{$^1$K. Singh and J. Leonard are with the Computer Science and Artificial Intelligence Laboratory (CSAIL) at the Massachusetts Institute of Technology (MIT), 32 Vassar St, Cambridge, MA 02139, USA.
        Corresponding author: Kurran Singh ({\tt\small singhk@mit.edu})}%
}

\begin{document}

\maketitle
\thispagestyle{empty}
\pagestyle{empty}

\begin{abstract}
Underwater object-level mapping requires incorporating visual foundation models to handle the uncommon and often previously unseen object classes encountered in marine scenarios. In this work, a metric of semantic uncertainty for open-set object detections produced by visual foundation models is calculated and then incorporated into an object-level uncertainty tracking framework. Object-level uncertainties and geometric relationships between objects are used to enable robust object-level loop closure detection for unknown object classes. The above loop closure detection problem is formulated as a graph-matching problem. While graph matching, in general, is NP-Complete, a solver for an equivalent formulation of the proposed graph matching problem as a graph editing problem is tested on multiple challenging underwater scenes. Results for this solver as well as three other solvers demonstrate that the proposed methods are feasible for real-time use in marine environments for the robust, open-set, multi-object, semantic-uncertainty-aware loop closure detection. Further experimental results on the KITTI dataset demonstrate that the method generalizes to large-scale terrestrial scenes.

\end{abstract}


\section{Introduction}
Achieving object-based mapping is a critical step in building more capable autonomous underwater systems, as it can enable higher level autonomous behaviors, while also providing for compressed map representations in low-bandwidth environments as well as human-interpretable maps for diver-AUV teaming or human-in-the-loop systems. Making such a system open-set, in that it should be able to identify and classify objects that were not in the training set, is also an integral aspect of underwater object-based mapping, since many underwater objects are not identified by current closed-set object detectors without fine tuning on hand labeled data, of which there is very little publicly available. 

Previous work \cite{singh2024opensetslam} \cite{singh2024optiacousticsemanticslamunknown} has demonstrated that latent vectors produced by clustering the output of visual foundation models can be useful representations of objects for data association and simultaneous localization and mapping (SLAM), especially when used alongside metrics of geometric uncertainty. However, the above works do not account for semantic uncertainty or the relative location of multiple objects and instead associate individual measurements with individual landmarks. As argued in \cite{frey2019efficient}, it is often safer to postpone loop closures and simply continue to build local maps until multiple object correspondences can be found, allowing for a more confident loop closure. Such methods would be useful for multi-robot map merging, multi-session relocalization, as well as for more traditional single vehicle loop closure detection. In this work, the problem of open-set object-based loop closures is formulated as a graph matching problem, thus enabling the use of the relative layout of multiple objects, represented by their latent vectors, for place recognition. Furthermore, notions of \textit{semantic} uncertainty are taken into account in the graph matching formulation through a method for the calculation and use of an uncertainty score for the output of visual foundation models. While in general, graph matching is NP-complete \cite{garey1979computers}, this work demonstrates that it is computationally feasible to solve the proposed problem formulation in realistic SLAM scenarios quickly and accurately using multiple different graph matching solver techniques. The proposed formulation is tested on data collected in underwater scenarios, thus demonstrating the use of the proposed framework in enabling robust, semantic-uncertainty-aware, open-set place recognition and map merging. 

Thus, the contributions of this paper are as follows:

\begin{enumerate}
    \item A method for calculating and tracking the uncertainty of open-set semantic encodings for underwater objects 
    \item A formulation of semantic-uncertainty-aware open-set object level place recognition as a graph matching problem that can be solved efficiently and accurately
    \item Experimental results demonstrating that the proposed methods enable real-time accurate place recognition in challenging underwater scenarios. Experiments are also performed on a large scale autonomous driving dataset, demonstrating that the proposed methods generalize to settings other than marine environments. 
\end{enumerate}

\section{Related Work}
\begin{figure*}[h] 
\center
  \includegraphics[width=\linewidth]{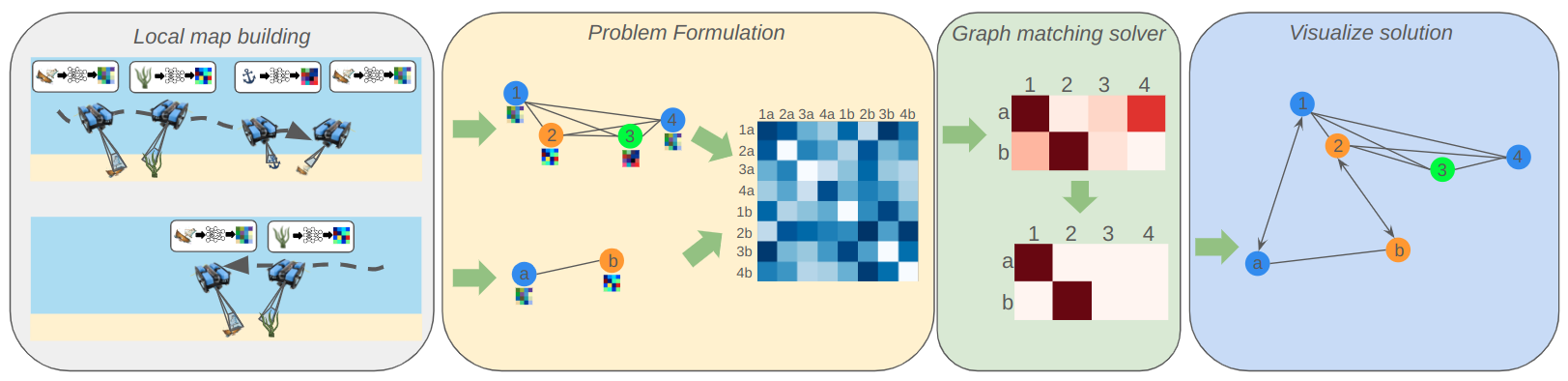}
  \caption{An overview of the proposed approach: In the initial local map building phase (top row), a vehicle navigates through an underwater scene and constructs a local map of the environment using latent vectors to represent objects detected by an open-set detector and localized using sonar information. Later (bottom row), the vehicle sees part of the same scene as in the initial pass, and creates a local map that has overlap with the initial scene. Next, in the problem formulation phase, the two local maps are converted into graphs where the nodes are uncertainty weighted latent vectors for each object, and the edges are the relative distances between the objects. From these graphs, an affinity matrix is created, using the node and edge affinity functions defined in \ref{sec:formulation}. By using a graph matching solver, a soft matching matrix is produced. The soft matching matrix is used to then obtain a hard matching matrix via the Hungarian algorithm. Finally, the hard match is used to visualize the object-level correspondences. }
  \label{fig:sys_overview}
\end{figure*}  

In \cite{singh2024optiacousticsemanticslamunknown}, it is demonstrated that object-based mapping is more accurate in underwater scenarios than feature-based methods due to lighting effects. Furthermore, in this paper, we argue that performing data association across multiple objects is a more reliable way to perform data association. Below, we provide a review of existing works on multi-object-based place recognition, including methods that formulate the problem as a graph matching problem.  

\subsection{Object-Based Place Recognition}
One of the earliest works to explore object-based place recognition as a graph matching problem is by Finman et al.~\cite{Finman15icraws}, which uses the object's discrete class and position to build the graph, which they solve using a brute force exhaustive search. 
Kong et al.~\cite{9341060} present a point cloud based method that extracts objects from KITTI using a closed-set detector. They then implement a method to learn correspondences between graphs created from the extracted objects. 
Ma et al.~\cite{10414178} also demonstrate a method that extracts a small number of classes of objects along with various geometric descriptors from a point cloud for use in matching, without accounting for semantic uncertainty in the vertex affinity function. There is no edge affinity taken into account in their formulation. 
Liu et al.~\cite{8794475} have a camera-based approach tested across various lighting changes that incorporates a semantic descriptor of the objects as well as the location of the objects into a topological map, but they then localize individual objects against the larger map rather than graph matching over multiple objects as in our approach. 
Yu et al.~\cite{9984819} formulate a camera-based approach to finding loop closures between graphs of everyday objects seen in the TUM and SceneNN datasets. They use a confusion matrix to account for semantic uncertainty, which is not possible to use for our open-set scenario. They also only test using the spectral matching solver as opposed to our work which experiments with multiple graph matching solvers.  
Gao et al.~\cite{10341898} propose a camera-based object graph matching approach that takes into account the semantics of the object as well as a learned set of geometric descriptors. 
In GOReloc, Wang et al.~\cite{10634741} perform camera-based graph matching at the object level that takes into account semantic uncertainty for closed-set object detections. 

A further work in the line of using the object-level geometric descriptors is \cite{10460989}, in which Cao et al. create a topological map of object-level quadrics which they then can localize against using 2D semantic features. 
Ji et al.~\cite{ji2023loopclosuredetectionbased} define their nodes to contain the object semantic label, position, color, and embedding, with edges representing spatial distance between objects. However, their method also is only for closed-set detectors, and furthermore does not take into semantic uncertainty.  

In \cite{frey2019efficient}, Frey et al. merge semantic maps through what they refer to as constellations of objects, which is conceptually similar to what we refer to as local object submaps. However, their focus is on the efficient extraction of the uncertainty-aware geometric distances between objects, rather than on the semantics. Similarly, Xing et al. perform map-merging based on learned descriptors of the constellations in \cite{Xing2022}. 

Lusk et al.~\cite{clipper} describe a relaxation of the graph matching problem that they call CLIPPER that can be efficiently and accurately solved for a variety of scenarios. 
In a different application, Fernandez-Cortizas et al.~\cite{10529513} demonstrate that semantic objects in the form of a scene graph can be used for multi-robot collaborative SLAM. Another application is for aerial vehicles localizing against trees across multiple seasons in \cite{thomas2024sosmatchsegmentationopensetrobust}. However, their work does not include the semantic embeddings of the objects as part of the nodes' information. 
For more information on the mechanics of solving graph matching problems, \cite{yanicmr16} by Yan et al. provides a survey of graph matching techniques. 

Our method differs from all of the above works in that it 1) allows for open-set object detections and the corresponding semantic uncertainties, 2) proposes multiple different node affinity functions to experimentally compare, and 3) tests several different solvers on the proposed formulation, with results demonstrating that the choice of both node affinity function and graph matching solver affects the accuracy of the results. Furthermore, we provide experimental results demonstrating that our method can work in difficult underwater scenes.

\section{Method}
\subsection{Uncertainty Quantification}\label{sec:uncertainty}
While closed-set semantic SLAM methods have well developed theory around semantic uncertainty through the use of a confusion matrix \cite{Doherty2019}, such methods are inapplicable to open-set scenarios due to the possibility of infinite object classes. Thus, new techniques need to be developed for calculating and incorporating uncertainties from foundation models. 
An important consideration in developing such an uncertainty model, especially for underwater scenarios, is the need for calculating aleatoric uncertainty rather than epistemic. This is due to the nature of the fact that nearly all underwater images and marine objects will be from outside of the training distribution, and therefore will all be associated with high epistemic uncertainty. Thus methods that simply detect whether an input is out-of-distribution from the training data, such as \cite{sharma2021sketching}, are insufficient. Rather, it is critical to be able to calculate and incorporate aleatoric uncertainty such that even when all objects in a scene are from outside of the training distribution, the uncertainty metric is able to provide \textit{relative} uncertainties with respect to the other images and objects. 
Therefore, the method of Kirchof et al~\cite{kirchhof2024pretrained} is leveraged to provide uncertainties. Their method is demonstrated in \cite{kirchhof2023url} to be independent of epistemic uncertainty, and instead an indicator of aleatoric uncertainty on a wide variety of datasets, given training data from ImageNet \cite{deng2009imagenet}.
Given an existing trained network, the uncertainty metric is produced through a multi-layer perceptron that attempts to predict the loss of the pretrained network on a given network. Furthermore, the uncertainty prediction network loss function is modified to incorporate ranking inputs by uncertainty as in \cite {yoo2019learning} to generate relative uncertainties between inputs while also untying the network output's scale from the training dataset.

\begin{equation}
\begin{aligned}
& \mathcal{L}=\max \left(0, \mathbb{1}_{\mathcal{L}} \cdot\left(u\left(e\left(x_1\right)\right)-u\left(e\left(x_2\right)\right)+m\right)\right)\\
& \text { s.t. } \\
& \mathbb{1}_{\mathcal{L}}:=\left\{\begin{array}{l}
+1, \text { if } \mathcal{L}_{\text {task }}\left(y_1, f\left(x_1\right)\right)>\mathcal{L}_{\text {task }}\left(y_2, f\left(x_2\right)\right) \\
-1, \text { else }
\end{array}\right.
\end{aligned}
\end{equation}

\subsection{Graph Matching Formulation}\label{sec:formulation}
We introduce a formulation of graph matching for object-based place recognition as a form of the Quadratic Assignment Problem (QAP) first proposed by Lawler \cite{lawler_qap}: 

\begin{equation}
\begin{aligned}
&\max_{\mathbf{X}} \ \text{vec}(\mathbf{X})^\top \mathbf{K} \text{vec}(\mathbf{X})\\
s.t. \quad &\mathbf{X} \in \{0, 1\}^{n_1\times n_2}, \ \mathbf{X}\mathbf{1} = \mathbf{1}, \ \mathbf{X}^\top\mathbf{1} \leq \mathbf{1}
\end{aligned}
\end{equation}

where $\mathbf{X}$ is the permutation matrix such that $\mathbf{X}_{i,a}=1$ indicates that node $i$ in graph 1 is matched to node $a$ in graph 2, and $\mathbf{X}_{i,a}=0$ means that there is no match. $\mathbf{K}$ is the affinity matrix that encodes node-wise and edge-wise affinities such that diagonal entry $\mathbf{K}_{i + a\times n_1, i + a\times n_1}$ is the node-wise affinity of node $i$ in graph 1 and node $a$ in graph 2, while off-diagonal entry $\mathbf{K}_{i + a\times n_1, j + b\times n_1}$ is the edge-wise affinity of edge $ij$ in graph 1 and edge $ab$ in graph 2.
The second and third constraints of the problem indicate respectively that the sum of each row must equal 1, and sum of each column must less than or equal to 1 i.e. assuming without loss of generality that graph 1 is a (candidate) subgraph of graph 2, graph 1 nodes each must be matched to exactly one node in graph 2, while graph 2 nodes can have no more than one match in graph 1, but may have no match at all.  

We define the nodes as 384-dimensional encodings produced by open-set object detection \cite{singh2024lossslamlightweightopensetsemantic}, representing a semantic encoding of each object via aggregation of clustered DINO \cite{Caron2021} features. Each node also stores uncertainty information as extracted in Section \ref{sec:uncertainty} and then tracked and updated as in Section \ref{sec:uncertainty_tracking}.



One proposed node affinity function is an uncertainty weighted cosine similarity: 

\begin{equation}
\text{WeightedCosineSim} = \frac{F_1 \cdot F_2}{\|F_1\| \|F_2\|} \times \frac{1}{1 + \frac{\sigma_1 + \sigma_2}{2}}
\end{equation}

Another proposed node affinity function is simply the Mahalanobis distance.

Finally, given the assumption that the embeddings display the noise characteristics of a multivariate Gaussian, we propose the use of the following closed form expression for the Bhattacharyya distance between two multivariate Gaussians as an uncertainty-aware node affinity function: 
\begin{equation}
\begin{aligned}
D_B\left(p_1, p_2\right) = & \frac{1}{8}\left(\boldsymbol{\mu}_1-\boldsymbol{\mu}_2\right)^T \boldsymbol{\Sigma}^{-1}\left(\boldsymbol{\mu}_1-\boldsymbol{\mu}_2\right)+\\
& \frac{1}{2} \ln \left(\frac{\operatorname{det} \boldsymbol{\Sigma}}{\sqrt{\operatorname{det} \boldsymbol{\Sigma}_1 \operatorname{det} \boldsymbol{\Sigma}_2}}\right)
\end{aligned}
\end{equation}

\noindent
where $ \Sigma = \frac{\Sigma_1 + \Sigma_2}{2}$ \cite{Kashyap_2019}. 

Note that this can be understood as the distance between the two distributions, such that e.g. two distributions with similar means and high, but similar, uncertainties will have a low Bhattacharyya distance. Thus, this metric can be understood intuitively as a way of incorporating the uncertainty as an additional attribute of the object itself, rather than as an more traditional uncertainty weighted distance as in the cosine similarity weighted by uncertainty, or the Mahalanobis distance. 

The edge affinity function in our proposed formulation is: 

\begin{equation}
\text{Affinity}(\mathbf{e}_{ab}, \mathbf{e}_{ij}) = \exp(-\frac{(\mathbf{e}_{ij} - \mathbf{e}_{ab})^2}{\sigma})
\end{equation}

\noindent
which is a Gaussian centered at the difference in Euclidean distance given that we define edge $ij$ as the Euclidean distance between node $i$ and node $j$. Alternative edge weights have been explored in other works and work exactly the same regardless of whether the scenario is open-set or closed-set, and as such, the focus of this work is to determine combinations of solvers and node affinity functions that work best for the open-set scenario. 

The solvers as described in Section \ref{sec:solvers} for this formulation output soft matching matrices. To obtain the final hard node correspondences, the soft matching matrix is fed into a linear solver utilizing the Hungarian algorithm \cite{hungarian} to obtain the final (hard) correspondences. A graphical overview of our entire approach can be seen in Figure \ref{fig:sys_overview}. 

\subsection{Graph Matching Solvers}\label{sec:solvers}
A variety of solvers for Lawler's QAP are tested. One solver is a spectral graph matching method \cite{leordeanuiccv05}, also known as the power iteration method. It computes the leading eigenvector of the affinity matrix $K$ by power iteration, and recovers assignments by using the principal eigenvector of $K$ to determine how strongly assignments belong the main cluster of $K$. 
Reweighted Random Walk Matching \cite{minsueccv2010} (RRWM) obtains a solution through simulating random walks with Sinkhorn reweighted jumps to enforce the constraints. 
The A* solver \cite{astar} finds the optimal matching between two graphs by solving the equivalent problem of the graph edit distance between two graphs, which is implemented in Pygmtools using a Hungarian heuristic. However, in our testing, we use a beam width of 0, which guarantees that the algorithm will find an optimal solution, albeit at the cost of greater computational complexity. Nevertheless, we find that the solver is still quick for our formulation on the tested underwater scenarios, as described in Section \ref{sec:discussion}. 
Finally, we test a neural solver \cite{WangPAMI22}, which attempts to learn correspondences between graphs. 
We use the Pygmtools library \cite{wang2024pygm} implementations of the solvers. 

\subsection{Local Map Construction} 
Local maps are constructed following the method of opti-acoustic fusion in \cite{singh2024optiacousticsemanticslamunknown}, which is reviewed here in brief. Odometry is provided by an Extended Kalman Filter fusing a pressure sensor, Doppler velocity log (DVL), and inertial measurement unit (IMU). Objects are identified using clustering on DINO features \cite{Caron2021} that are produced through a visual transformer network \cite{Dosovitskiy2020}. The objects are localized using opti-acoustic fusion of the camera and sonar data, and are represented by the latent space centroid of all of the object's features. Associations between a new observation and existing landmarks are made using both a cosine similarity test

\begin{equation}
\text{Cos Sim}(\mathbf{n}_a, \mathbf{n}_i):=\dfrac {\mathbf{n}_a \cdot \mathbf{n}_i} {\left\| \mathbf{n}_a\right\| _{2}\left\| \mathbf{n}_i\right\| _{2}} 
\end{equation}




\noindent
between the semantic encodings and a Mahalanobis distance check. Furthermore, in a slight modification from the original method, a temporal check is added to ensure that the maps are indeed local, and that global loop closures are not made, with the exception of run 1 as described in Section \ref{sec:experiments}, which allows for loop closures in order to simulate a multi-session mapping scenario rather than a pure loop closure detection scenario.  

If no landmarks meet the three criteria (cosine similarity, Mahalanobis distance, temporal) above, then a new landmark is added. Otherwise, the max-likelihood hypothesis is selected given prior estimates $\poses^0, \landmarks^0$. 
\begin{equation}
\hat{\associations}=\underset{\associations}{\arg \max } \; p\left(\associations \mid \poses^0, \landmarks^0, \measurements\right)
\end{equation}
where $\associations \triangleq \{\association_k : \association_k \in \mathbb{N}_{\leq M},\ k = 1, \ldots K\}$ and $M$ is the number of existing landmarks. 
Given the selected data association, the pose and landmark estimates are then updated: 
\begin{equation}
\hat{\poses}, \hat{\landmarks}=\underset{\poses \in SE(3), \landmarks \in \mathbb{R}^{3}}{\arg \max } \; \log p(\measurements \mid \poses, \landmarks, \hat{\associations}).
\end{equation}
The poses, landmarks, odometry measurements, and landmark measurements are formulated as a factor graph, which then can be solved efficiently using iSAM2 \cite{Kaess12ijrr}.

\subsection{Object Uncertainty Tracking}\label{sec:uncertainty_tracking}

\begin{figure}
    \vspace{2mm}
     \centering
     \begin{subfigure}[b]{0.32\linewidth}
     \center
         \includegraphics[width=.9\linewidth]{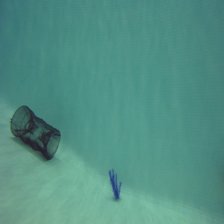}
         \label{fig:cage_and_seaweed}
     \end{subfigure}
     \hfill
     \begin{subfigure}[b]{0.32\linewidth}
         \centering
         \includegraphics[width=.9\linewidth]{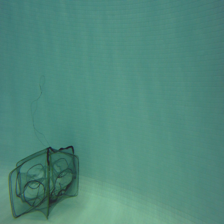}
         \label{fig:cage}
     \end{subfigure}
     \hfill
     \begin{subfigure}[b]{0.32\linewidth}
     \center
         \includegraphics[width=0.9\linewidth]{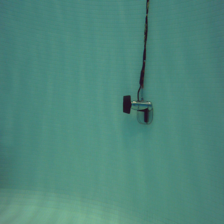}
         \label{fig:towfish}
     \end{subfigure}
        \caption{Images with low ($< 0.15$) uncertainty.}
        \label{fig:low_uncertainties}
\end{figure}

\begin{figure}
     \centering
     \begin{subfigure}[b]{0.32\linewidth}
     \center
         \includegraphics[width=.9\linewidth]{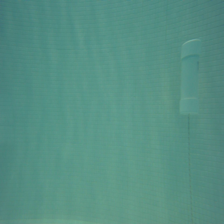}
         \label{fig:buoy}
     \end{subfigure}
     \hfill
     \begin{subfigure}[b]{0.32\linewidth}
         \centering
         \includegraphics[width=.9\linewidth]{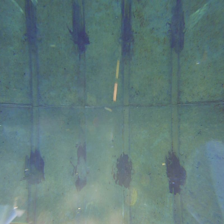}
         \label{fig:outdoor_wall}
     \end{subfigure}
     \hfill
     \begin{subfigure}[b]{0.32\linewidth}
     \center
         \includegraphics[width=0.9\linewidth]{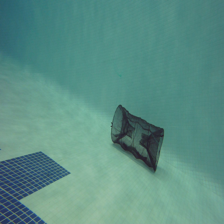}
         \label{fig:cage_with_tiles}
     \end{subfigure}
        \caption{Images with high ($> 0.35$) uncertainty.}
        \label{fig:high_uncertainties}
\end{figure}

Given a data association decision that an observation has been made of a landmark, the semantic component of the landmark $l_{sem}$ along with its uncertainty is tracked as follows. Note that uncertainties are extracted at the image level, whereas tracking is done at the object level i.e. we update the uncertainties for each object in an image with the image-level uncertainty. 

A prior distribution of $l_{sem}$ is given by the first observation of a new object: 

\begin{equation}
    l_{sem} \sim \mathcal{N}(\mu_0, \Sigma_0)
\end{equation}

\noindent where $\mu_0$ is the prior mean and $\Sigma_0$ is the prior covariance matrix. 

The measurement model is as follows:

\begin{equation}
    y_i \sim \mathcal{N}(x, \Sigma_i)
\end{equation}

\noindent where each measurement $y_i$ is a directy noisy observation of $x$, and $\Sigma_i$ is the covariance matrix of the measurement noise for the i-th observation.

The posterior update can then be performed using the Kalman filter equations. The innovation covariance $S_i$ is given by:

\begin{equation}
    S_i = \Sigma_{\text{post},i-1} + \Sigma_i
\end{equation}

\noindent where $\Sigma_{\text{post},i-1}$ is the covariance of the posterior distribution after the $(i-1)$-th update. The Kalman gain $K_i$ is then

\begin{equation}
    K_i = \Sigma_{\text{post},i-1} S_i^{-1}
\end{equation}

\noindent Thus the posterior mean and posterior covariance after incorporating the $i$-th measurement are updated respectively as:

\begin{subequations}
\begin{equation}
\mu_{\text{post},i} = \mu_{\text{post},i-1} + K_i \left( y_i - \mu_{\text{post},i-1} \right)
\end{equation} 
\begin{equation}
\Sigma_{\text{post},i} = \Sigma_{\text{post},i-1} - K_i S_i K_i^\top.
\end{equation}
\end{subequations}

\noindent Finally, the posterior distribution of $l_{sem}$ after all measurements have been incorporated is: 

\begin{equation}
x \mid \{y_i\}_{i=1}^n \sim \mathcal{N}(\mu_{\text{post},n}, \Sigma_{\text{post},n})
\end{equation}
where $\mu_{\text{post},n}$ and $\Sigma_{\text{post},n}$ are the mean and covariance of $l_{sem}$ after all updates. 

While several operations in this above framework are computationally expensive, by leveraging the fact that the noise matrices are diagonal, inverting the innovation covariance can be completed through simply taking an element-wise reciprocal rather than LU decomposition, while all matrix multiplications can also be performed more quickly than in the general case.

\section{Experimental Evaluation}\label{sec:experiments}
The proposed system is tested on a public underwater dataset \cite{singh2024optiacousticsemanticslamunknown} that contains challenging lighting conditions and uncommon objects such as towfish, lobster cages, and buoys. The trajectories are split at the halfway point, with different subsections of the second half used as smaller local maps to be matched to the original map from the first lap. Run 1 consists of four laps in total, thus making for two separate two lap trajectories, as opposed to run 2 and run 3, which are two laps in total, and thus each half only allows the vehicle to make a single pass across the mapping scene. Note that the run numbers do not correspond directly with the run numbers in \cite{singh2024optiacousticsemanticslamunknown}, as we only evaluate on three of their four datasets. Edges are added between objects if the Euclidean distance between their centers is less than 2 meters, with this constraint relaxed as needed to ensure a connected graph. 

Evaluations are also run on the KITTI dataset, with four 20 node subgraphs extracted at the main loop closure opportunities from the full graph creates over 2.2 kilometers of driving. The subgraphs are then matched to the full graph which consists of 605 objects from 14 classes extracted by the same open-set object detector as with the underwater datasets. Depth is obtained via stereo depth, and stereo odometry is obtained from RTABMAP. Edges are added between all objects within 100 meters of each other. 
\begin{figure}[] 
\center
  \includegraphics[width=\linewidth]{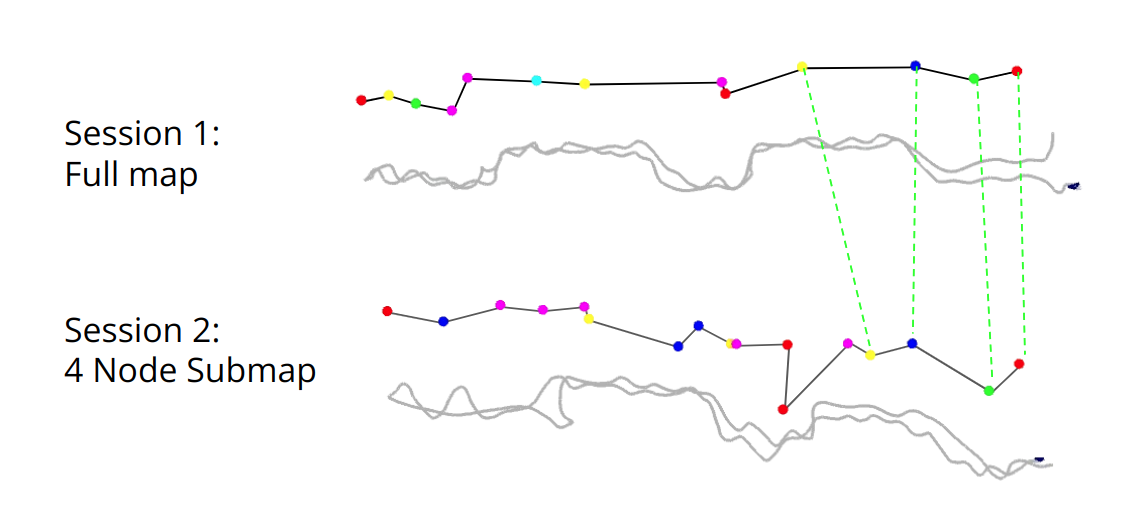}
  \caption{Matching results from run 1. The method is able to find the correct correspondences for the 4 node submap in the original full map despite noisy object locations due to noise from sonar-based ranging, as well as handling the object detection uncertainties from detecting objects with underwater lighting effects. }
  \label{fig:2obj2loop_correspondences}
\end{figure}





\begin{figure}[] 
\center
  \includegraphics[width=\linewidth]{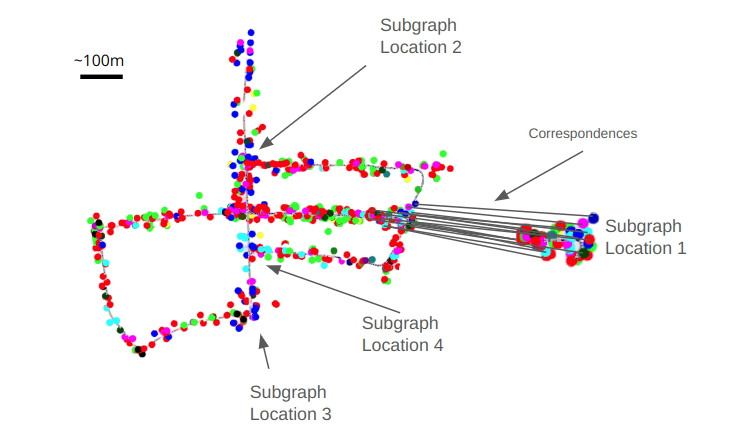}
  \caption{Results on the KITTI dataset demonstrate that the technique is generalizable to terrestrial settings, while also being feasible for larger scale graph matching. The different colors represent different object classes. Subgraphs were extracted at the following locations where loop closure opportunities arise in the trajectory. The correspondences are drawn for illustrative purposes. }
  \label{fig:kitti_correspondences}
\end{figure}

\section{Discussion}\label{sec:discussion}
The proposed method for obtaining uncertainties for visual foundation based object extraction works well for underwater scenes, as the images with low uncertainty tend to be of clearly visible objects with no background texture as in Figure \ref{fig:low_uncertainties}, whereas images with high uncertainty tend to be of objects with textured backgrounds, difficult lighting effects, or objects with low contrast with the background as in Figure \ref{fig:high_uncertainties}. 

\begin{table}[b!]
\begin{center}
\caption{KITTI Results. See Figure \ref{fig:kitti_correspondences} for the location labels. At each loop closure opportunity, a 20 object subgraph is extracted to match against the full graph. -- indicates the solver was halted after 5 minutes with no solution.}
\begin{tabular}{c c c c c c c } 
 \hline
 & & \multicolumn{4}{c}{Location} \\
 \hline
 Solver & Affinity & 1 & 2 & 3 & 4 & Avg. \\ [0.1ex] 
 \hline 
 A* & Unc. Cos & -- & -- & -- & -- & --\\ 
       & Bhatt. & -- & -- & -- & -- & --\\
       & Mah. & -- & -- & -- & -- & -- \\  \hline 

 RRWM & Unc. Cos & 0.95 & 0.52 & 0.70 & 0.80 & 0.74 \\
      & Bhatt. & 0.95 & 0.57 & 0.50 & 0.85 & 0.72 \\
      & Mah. & 0.90 & 0.57 & 0.50 & 0.70 & 0.67 \\ \hline 

Spectral & Unc. Cos & 1.00 & 0.00 & 0.00 & 0.00 & 0.25       \\
                  & Bhatt. & 0.95 & 0.00 & 0.00 & 0.00 & 0.24      \\
                  & Mah. & 0.81 & 0.00 & 0.00 & 0.00 & 0.20      \\ \hline 
Neural & Uncertainty Cos & 0.00 & 0.38 & 0.65 & 0.45 & 0.37     \\
                  & Bhatt. & 0.00 & 0.29 & 0.45 & 0.40 & 0.29      \\
                  & Mah. & 0.00 & 0.10 & 0.25 & 0.35 & 0.18      \\ \hline 
\end{tabular}

\label{table:kitti}
\end{center}
\end{table}

For the underwater experiments, Figure \ref{fig:2obj2loop_correspondences} demonstrates that the local maps themselves can be noisy and full of false positives due to difficult lighting effects and erroneous ranges from acoustic effects seen in the sonar data. However, despite the noise present in the maps, it is possible for the A*, RRWM, and SM solvers to obtain up to 100\% accuracy when the weighted cosine function is used for node affinities. The neural graph solver exhibits steadier performance across the smaller numbers of nodes in the subgraph, and the neural solver in general does not follow the same accuracy trends as the traditional solvers. This is likely due to the neural solver's training on a particular set of graph matching problems that are not necessarily aimed at generalizing as well across a wide variety of graph matching problems. 
Generally, the solvers (other than the neural solver) perform better with more nodes, thus proving the notion that it is important to use this method rather than simply trying to align individual objects. All solvers are fast enough to use in realistic robotics scenarios, with the creation of adjacency matrices taking 0.008 seconds, the creation of the affinity matrices taking 

\begin{figure*}[h!]
    \vspace{2mm}
\end{figure*}
\begin{table*}[]
\begin{center}
\caption{Graph matching accuracy for the underwater datasets. Accuracy figures are at the node level averaged across 2, 3, 4, and 5 node subgraphs matched against the full map during loop closure opportunities when odometric drift is at its highest.}
\vspace{.75cm}
\begin{tikzpicture}[remember picture, overlay]
    \node at (-0.5, 1.2) {};  
    \draw[decoration={brace,amplitude=5pt},decorate,thick] (-5.8,0.1) -- (-1.6,0.1);
    \node at (-3.7, 0.5) {Run 1};
    \node at (-0.5, 1.2) {};  
    \draw[decoration={brace,amplitude=5pt},decorate,thick] (-1.2,0.1) -- (3.0,0.1);
    \node at (0.8, 0.5) {Run 2};
    \node at (-0.5, 1.2) {};  
    \draw[decoration={brace,amplitude=5pt},decorate,thick] (3.4,0.1) -- (7.6,0.1);
    \node at (5.5, 0.5) {Run 3};
    \node at (-0.5, 1.2) {};  
    \draw[decoration={brace,amplitude=5pt},decorate,thick] (7.8,0.1) -- (8.7,0.1);
    \node at (8.1, 0.5) {Aggregate};
\end{tikzpicture}

\begin{tabular}{c c c c c c c c c c c c c c c c c c} 
 \hline
 & & \multicolumn{4}{c}{Subgraph Size} & & \multicolumn{4}{c}{Subgraph Size} & & \multicolumn{4}{c}{Subgraph Size} \\ 
 \hline
 Solver & Affinity & 2 & 3 & 4 & 5 & \textit{Avg.} & 2 & 3 & 4 & 5 & \textit{Avg.} & 2 & 3 & 4 & 5 & \textit{Avg.} & \textit{Avg.} \\ [0.1ex] 
 \hline 
 A*   & Unc. Cos & 1.00 & 1.00 & 1.00 & 1.00 & \textit{1.00} & 1.00 & 1.00 & 1.00 & 0.80 & \textit{0.95} & 0.50 & 0.67 & 0.50 & 1.00 & \textit{0.67} & \textbf{\textit{0.83}}\\ 
       & Bhatt. & 0.00 & 0.00 & 0.00 & 1.00 & \textit{0.25} & 0.00 & 0.00 & 0.00 & 0.00 & \textit{0.00} & 0.50 & 0.00 & 0.00 & 0.00 & \textit{0.13} & \textit{0.13}\\
       & Mah. & 0.00 & 0.00 & 0.00 & 0.00 & \textit{0.00} & 0.00 & 0.00 & 0.00 & 0.00 & \textit{0.00} & 0.50 & 0.00 & 0.00 & 0.00 & \textit{0.13} & \textit{0.04}\\ 
       \hline 

 RRWM & Unc. Cos & 1.00 & 1.00 & 1.00 & 1.00 & \textit{1.00} & 1.00 & 1.00 & 1.00 & 0.60 & \textit{0.90} & 0.00 & 0.33 & 0.50 & 1.00 & \textit{0.46} & \textit{0.79}\\
      & Bhatt.  & 0.00 & 0.00 & 0.00 & 0.00 & \textit{0.00} & 0.00 & 0.00 & 0.00 & 0.00 & \textit{0.00} & 0.00 & 0.00 & 0.00 & 0.00 & \textit{0.00} & \textit{0.00}\\
      & Mah.  & 0.00 & 0.00 & 0.00 & 0.00 & \textit{0.00} & 0.00 & 0.00 & 0.00 & 0.00 & \textit{0.00} & 0.50 & 0.00 & 0.00 & 0.00 & \textit{0.13} & \textit{0.04}\\ 
      \hline 

Spectral & Unc. Cos & 1.00 & 1.00 & 1.00 & 1.00 & \textit{1.00} & 1.00 & 1.00 & 0.50 & 0.40 & \textit{0.73} & 0.00 & 0.33 & 0.25 & 1.00 & \textit{0.40} & \textit{0.71}\\
                  & Bhatt. & 0.00 & 0.00 & 0.00 & 0.60 & \textit{0.15} & 0.50 & 0.00 & 0.00 & 0.20 & \textit{0.18} & 0.50 & 0.33 & 0.00 & 0.20 & \textit{0.26} & \textit{0.20}\\
                  & Mah. & 0.00 & 0.00 & 0.00 & 0.40 & \textit{0.10} & 0.50 & 0.00 & 0.00 & 0.00 & \textit{0.13} & 0.50 & 0.33 & 0.00 & 0.20 & \textit{0.26} & \textit{0.16}\\ 
                  \hline 
Neural & Unc. Cos & 0.50 & 0.33 & 0.50 & 0.40 & \textit{0.43} & 0.50 & 0.33 & 0.25 & 0.20 & \textit{0.32} & 1.00 & 1.00 & 0.75 & 0.40 & \textit{0.79} & \textit{0.51} \\
                  & Bhatt. & 0.50 & 0.33 & 0.00 & 0.00 & \textit{0.21} & 0.00 & 0.00 & 0.25 & 0.00 & \textit{0.06} & 0.50 & 0.00 & 0.00 & 0.20 & \textit{0.18} & \textit{0.15}\\
                  & Mah.   & 0.50 & 0.33 & 0.00 & 0.00 & \textit{0.21} & 0.00 & 0.00 & 0.00 & 0.00 & \textit{0.00} & 0.00 & 0.00 & 0.00 & 0.20 & \textit{0.05} & \textit{0.09}\\ 
                  \hline 
\end{tabular}

\label{table:2obj2loop}
\end{center}
\end{table*}

\noindent 0.006 seconds, and the solvers taking 0.02 seconds on average for all numbers of nodes in the scenarios from table \ref{table:2obj2loop}. 
It is interesting to note that the Mahalanobis distance works poorly regardless of solver, despite perhaps being the most intuitive metric to use. However, the encoding latent space likely breaks the equal dispersion assumption, thus making highlighting the need to build more tailored node affinity functions such as the uncertainty weighted cosine distance. 

For the KITTI experiments, A* fails to scale to the larger sizes, and execution was halted after several minutes. However, the other solvers exhibit much stronger performance and provide reliably accurate results with reasonable speed. In particular, RRWM with both the uncertainty weighted cosine affinity function and the Bhattacharyya affinity function has the most accurate performance across the four loop closure opportunities. The neural solver is again the exception, in that it completely fails to adapt to these graphs, and most likely needs to be trained on more similar graph formulations to be more effective. The reason the solvers that were less accurate in the underwater scenario are more successful in the terrestrial setting is likely due to the much cleaner data, as stereo depth is more accurate than sonar-based ranging, and there are less challenging lighting effects than in the underwater scenes. While runtimes even for the RRWM and Spectral solvers increase to 0.0-0.5s for adjacency matrix creation, 0.5-0.8s for affinity matrix creation, and 1-2s for solves, such times are reasonable to use in a real world real-time scenario, since these loop closures are only for large scale, occasional use. 

The results demonstrate the importance of carefully choosing the node affinity function and solver for the application, with the accuracy figures showing that for open-set place recognition, larger subgraphs improve performance and that the weighted cosine distance function is the best choice for affinity function. While A* is the most accurate solver, it comes at the price of computational efficiency, and in larger scale graphs, RRWM achieves strong accuracy while also providing fast solutions. 

\section{Conclusion}
This work addresses the problem of semantic-uncertainty-aware graph matching for loop closure detection in challenging underwater environments. A method is proposed for obtaining and tracking uncertainties for the object detections and encodings produced by visual foundation models for open-set object detection. Various uncertainty-aware affinity functions are proposed as part of the graph matching problem formulation. While graph matching is NP-complete, multiple existing solvers are demonstrated to be feasible for real-time use on the proposed formulation. A method for obtaining the local maps is proposed and used on multiple underwater datasets. The produced local maps are then used to test the ability of the graph matching system to find correspondences across different local maps. The A* solver along with the uncertainty weighted cosine similarity function exhibits the highest accuracy in matching sections of local maps known as subgraphs to a larger local map, while RRWM along with the uncertainty weighted cosine similarity function achieves a good mix of accuracy and speed. 

\section*{Acknowledgments}
\footnotesize This work was supported by the MIT Lincoln Laboratory Autonomous Systems Line which is funded by the Under Secretary of Defense for Research and Engineering through Air Force Contract No. FA8702-15-D-0001, ONR grants
N00014-18-1-2832, N00014-23-12164, and N00014-19-1-2571 (Neuroautonomy MURI), and the MIT Portugal Program.

\bibliographystyle{IEEEtran}
\bibliography{main}

\end{document}